\documentclass[10pt,twocolumn,letterpaper]{article}

\usepackage{iccv}
\usepackage{times}
\usepackage{epsfig}
\usepackage{graphicx}
\usepackage{amsmath}
\usepackage{amssymb}

\usepackage[pagebackref=true,breaklinks=true,letterpaper=true,colorlinks,bookmarks=false]{hyperref}
\usepackage{bm}
\usepackage{booktabs}

\iccvfinalcopy 

\def\iccvPaperID{1201} 

\begin{document}

\title{Query-guided Regression Network with Context Policy for Phrase Grounding}
\author{
   Kan Chen\thanks{Equal contribution. Names are sorted alphabetically.}\qquad Rama Kovvuri\footnotemark[1]\qquad Ram Nevatia \\
   University of Southern California, Institute for Robotics and Intelligent Systems \\
   Los Angeles, CA 90089, USA \\
   {\tt\small\{kanchen|nkovvuri|nevatia\}@usc.edu}
}

\maketitle

\begin{abstract}
    Given a textual description of an image, phrase grounding localizes objects in the image referred by query phrases in the description.
    State-of-the-art methods address the problem by ranking a set of proposals based on the relevance to each query, which are limited by the performance of independent proposal generation systems and ignore useful cues from context in the description.
    In this paper, we adopt a spatial regression method to break the performance limit, and introduce reinforcement learning techniques to further leverage semantic context information.
    We propose a novel Query-guided Regression network with Context policy (QRC Net) which jointly learns a Proposal Generation Network (PGN), a Query-guided Regression Network (QRN) and a Context Policy Network (CPN).
    Experiments show QRC Net provides a significant improvement in accuracy on two popular datasets: Flickr30K Entities and Referit Game, with 14.25\% and 17.14\% increase over the state-of-the-arts respectively.\vspace{-3.5mm}
\end{abstract}

\section{Introduction} 
Given an image and a related textual description, phrase grounding attempts to localize objects which are mentioned by corresponding phrases in the description. 
It is an important building block in computer vision with natural language interaction, which can be utilized in high-level tasks, such as image retrieval~\cite{Chen_2017_CVPR,radenovic2016cnn}, image captioning~\cite{karpathy2015deep,fang2015captions} and visual question answering~\cite{antol2015vqa,chen2015abc,fukui2016multimodal}.

\begin{figure}[t]
\includegraphics[width=3.2in]{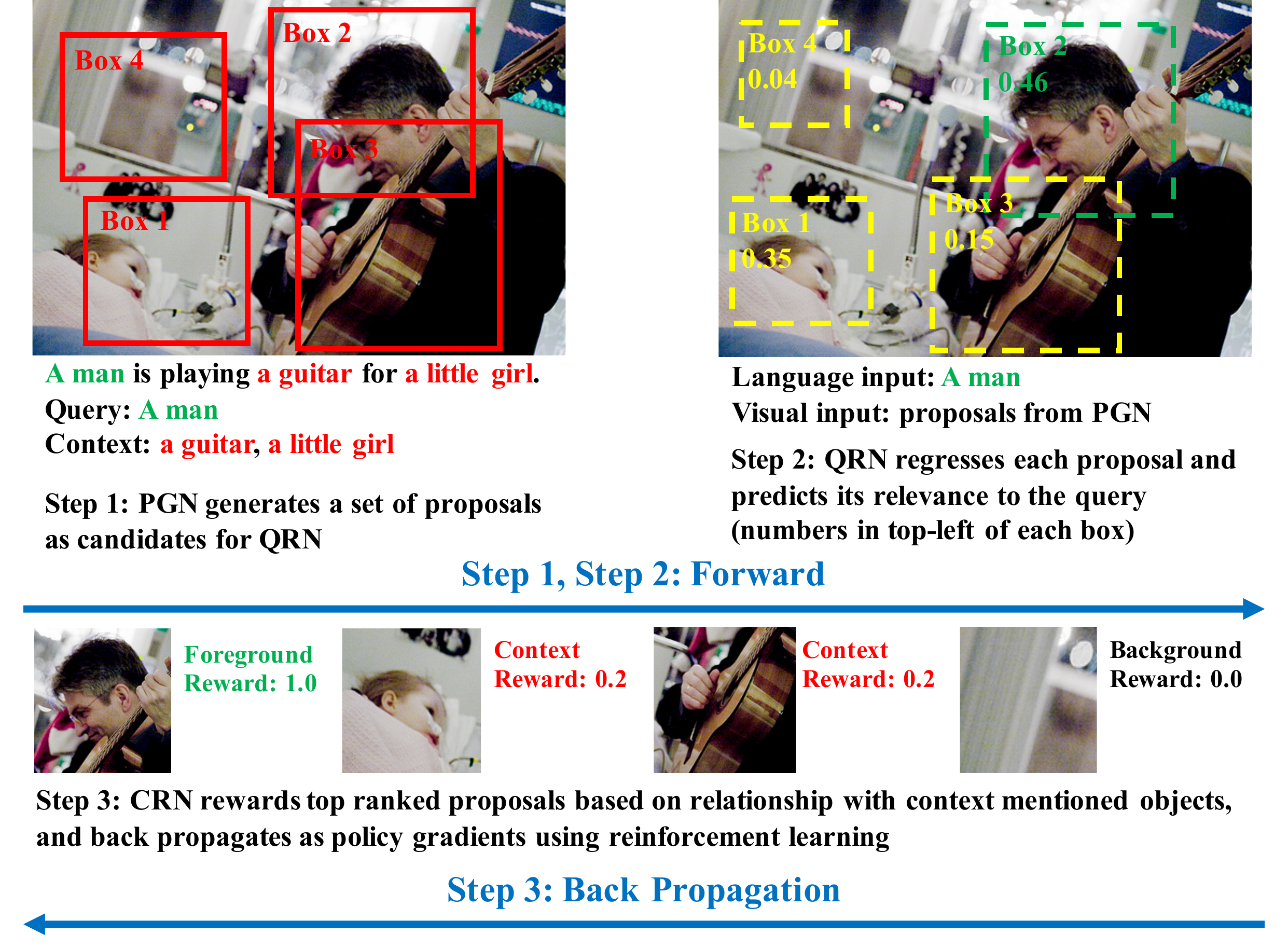}
\centering
\caption{QRC Net first regresses each proposal based on query's semantics and visual features, and then utilizes context information as rewards to refine grounding results.}\label{fig: intro}
\end{figure}

Phrase Grounding is a challenging problem that involves parsing language queries and relating the knowledge to localize objects in visual domain.
To address this problem, typically a proposal generation system is first applied to produce a set of proposals as grounding candidates. 
The main difficulties lie in how to learn the correlation between language (query) and visual (proposals) modalities, and how to localize objects based on multimodal correlation. 
State-of-the-art methods address the first difficulty by learning a subspace to measure the similarities between proposals and queries.
With the learned subspace, they treat the second difficulty as a retrieval problem, where proposals are ranked based on their relevance to the input query. 
Among these, Phrase-Region CCA~\cite{plummer2015flickr30k} and SCRC~\cite{hu2016natural} models learn a multimodal subspace via Canonical Correlation Analysis (CCA) and a Recurrent Neural Network (RNN) respectively. 
Varun \emph{et al.}~\cite{nagaraja2016modeling} learn multimodal correlation aided by context objects in visual content.
GroundeR~\cite{rohrbach2016grounding} introduces an attention mechanism that learns to attend on related proposals given different queries through phrase reconstruction.

These approaches have two important limitations. 
First, proposals generated by independent systems may not always cover all mentioned objects given various queries; since retrieval based methods localize objects by choosing one of these proposals, they are bounded by the performance limits from proposal generation systems.
Second, even though query phrases are often selected from image descriptions, context from these descriptions is not utilized to reduce semantic ambiguity.
Consider example in Fig~\ref{fig: intro}.
Given the query ``a man'', phrases ``a guitar'' and ``a little girl'' can be considered to provide context that proposals overlapping with ``a guitar'' or ``a little girl'' are less likely to be the ones containing ``a man''.

To address the aforementioned issues, we propose to predict mentioned object's location rather than selecting candidates from limited proposals.
We adopt a regression based method guided by input query's semantics. 
To reduce semantic ambiguity, we assume that different phrases in one sentence refer to different visual objects.
Given one query phrase, we evaluate predicted proposals and down-weight those which cover objects mentioned by other phrases (\emph{i.e.}, context). 
For example, we assign lower rewards for proposals containing ``a guitar'' and ``a little girl'' in Fig~\ref{fig: intro} to guide system to select more discriminative proposals containing ``a man''.
Since this procedure depends on prediction results and is non-differentiable, we utilize reinforcement learning~\cite{sutton1998reinforcement} to adaptively estimate these rewards conditioned on context information and jointly optimize the framework.

In implementation, we propose a novel Query-guided Regression network with Context policy (QRC Net) which consists of a Proposal Generation Network (PGN), a Query-guided Regression Network (QRN) and a Context Policy Network (CPN). 
PGN is a proposal generator which provides candidate proposals given an input image (red boxes in Fig.~\ref{fig: intro}).
To overcome performance limit from PGN, QRN not only estimates each proposal's relevance to the input query, but also predicts its regression parameters to the mentioned object conditioned on the query's intent (yellow and green boxes in Fig.~\ref{fig: intro}).
CPN samples QRN's prediction results and evaluates them by leveraging context information as a reward function. 
The estimated reward is then back propagated as policy gradients (Step 3 in Fig.~\ref{fig: intro}) to assist QRC Net's optimization.
In training stage, we jointly optimize PGN, QRN and CPN using an alternating method in~\cite{ren2015faster}. 
In test stage, we fix CPN and apply trained PGN and QRN to ground objects for different queries.

We evaluate QRC Net on two grounding datasets: Flickr30K Entities~\cite{plummer2015flickr30k} and Referit Game~\cite{KazemzadehOrdonezMattenBergEMNLP14}. 
Flickr30K Entities contains more than 30K images and 170K query phrases, while Referit Game has 19K images referred by 130K query phrases. 
Experiments show QRC Net outperforms state-of-the-art methods by a large margin on both two datasets, with more than 14\% increase on Flickr30K Entities and 17\% increase on Referit Game in accuracy.

Our contributions are twofold: First, we propose a query-guided regression network to overcome performance limits of independent proposal generation systems. Second, we introduce reinforcement learning to leverage context information to reduce semantic ambiguity. In the following paper, we first discuss related work in Sec.~\ref{sec: related work}. More details of QRC Net are provided in Sec.~\ref{sec: method}. Finally we analyze and compare QRC Net with other approaches in Sec.~\ref{sec: exps}.

\section{Related Work}\label{sec: related work}

\textbf{Phrase grounding} requires learning correlation between visual and language modalities.
Karpathy \emph{et al.}~\cite{karpathy2014deep} propose to align sentence fragments and image regions in a subspace, and later replace the dependency tree with a bi-directional RNN in~\cite{karpathy2015deep}.  
Hu \emph{et al.}~\cite{hu2016natural} propose a SCRC model which adopts a 2-layer LSTM to rank proposals using encoded query and visual features. 
Rohrbach \emph{et al.}~\cite{rohrbach2016grounding} employ a latent attention network conditioned on query which ranks proposals in unsupervised scenario. 
Other approaches learn the correlation between visual and language modalities based on Canonical Correlation Analysis (CCA)~\cite{hardoon2004canonical} methods. 
Plummer \emph{et al.}~\cite{plummer2014flickr30k} first propose a CCA model to learn the multimodal correlation. 
Wang \emph{et al.}~\cite{wang2016structured} employ structured matching and use phrase pairs to boost performance. 
Recently, Plummer \emph{et al.}~\cite{plummer2015flickr30k} augment the CCA model to leverage extensive linguistic cues in the phrases. 
All of the above approaches are reliant on external object proposal systems and hence, are bounded by their performance limits.

\textbf{Proposal generation and spatial regression.} 
Proposal generation systems are widely used in object detection and phrase grounding tasks.
Two popular methods: Selective Search~\cite{uijlings2013selective} and EdgeBoxes~\cite{zitnick2014edge} employ efficient low-level features to produce proposals on possible object locations.
Based on proposals, spatial regression method is successfully applied in object detection. 
Fast R-CNN~\cite{girshickICCV15fastrcnn} first employs a regression network to regress proposals generated by Selective Search~\cite{uijlings2013selective}. 
Based on this, Ren \emph{et al.}~\cite{ren2015faster} incorporate the proposal generation system by introducing a Region Proposal Network (RPN) which improves both accuracy and speed in object detection. 
Redmon \emph{et al.}~\cite{redmon2016you} employ regression method in grid level and use non-maximal suppression to improve the detection speed. 
Liu \emph{et al.}~\cite{liu2016ssd} integrate proposal generation into a single network and use outputs discretized over different ratios and scales of feature maps to further increase the performance.
Inspired by the success of RPN in object detection, we build a PGN and regress proposals conditioned on the input query.

\begin{figure*}[h]
\includegraphics[width=7.0in]{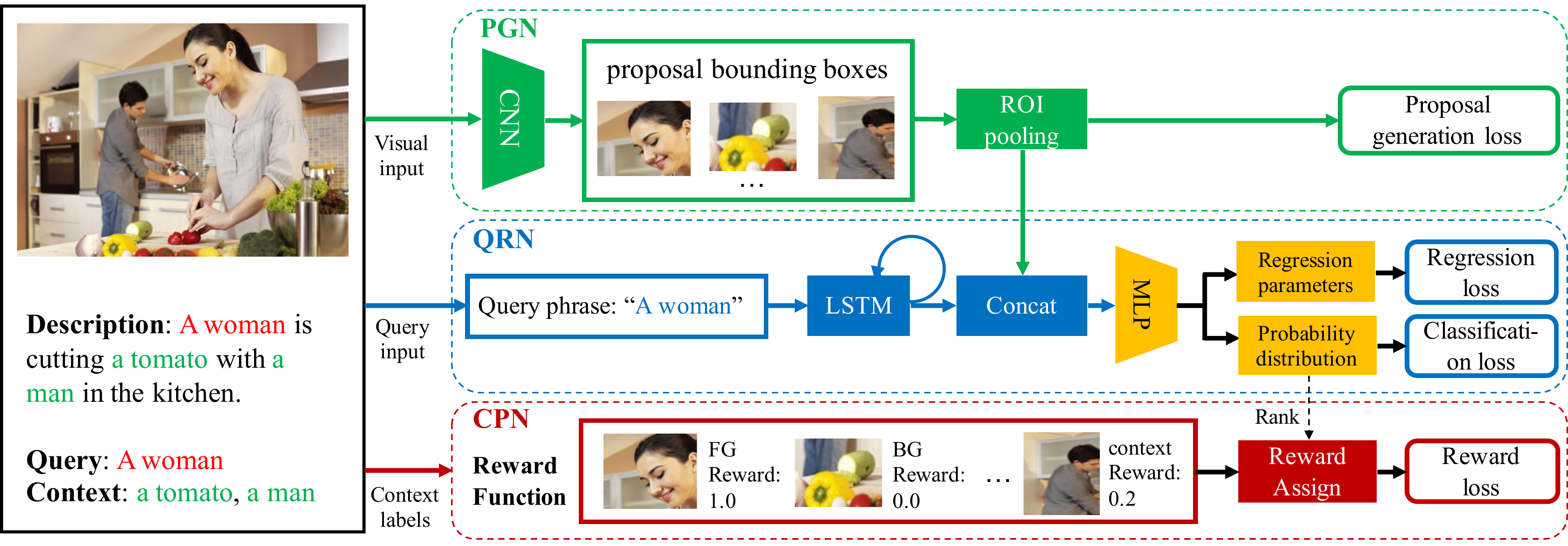}
\centering
\caption{Query-guided Regression network with Context policy (QRC Net) consists of a Proposal Generation Network (PGN), a Query-guided Regression Network (QRN) and a Context Policy Network (CPN). PGN generates proposals and extracts their CNN features via a RoI pooling operation~\cite{ren2015faster}. QRN encodes input query's semantics by an LSTM~\cite{hochreiter1997long} model and regresses proposals conditioned on the query. CPN samples the top ranked proposals, and assigns rewards considering whether they are foreground (FG), background (BG) or context. These rewards are back propagated as policy gradients to guide QRC Net to select more discriminative proposals.}\label{fig: QRC Net}
\end{figure*}

\textbf{Reinforcement learning} is first introduced to deep neural network in Deep Q-learning (DQN)~\cite{mnih2015human}, which teaches an agent to play ATARI games. 
Lillicrap \emph{et al.}~\cite{lillicrap2015continuous} modify DQN by introducing deep deterministic policy gradients, which enables reinforcement learning framework to be optimized in continuous space. 
Recently, Yu \emph{et al.}~\cite{yu2016joint} adopt a reinforcer to guide speaker-listener network to sample more discriminative expressions in referring tasks. 
Liang \emph{et al.}~\cite{liang2017deep} introduce reinforcement learning to traverse a directed semantic action graph to learn visual relationship and attributes of objects in images.
Inspired by the successful applications of reinforcement learning, we propose a CPN to assign rewards as policy gradients to leverage context information in training stage.

\section{QRC Network}\label{sec: method}

QRC Net is composed of three parts: a Proposal Generation Network (PGN) to generate candidate proposals, a Query-guided Regression Network (QRN) to regress and rank these candidates and a Context Policy Network (CPN) to further leverage context information to refine ranking results.
In many instances, an image is described by a sentence that contains multiple noun phrases which are used as grounding queries, one at a time. 
We consider the phrases that are not in the query to provide context; specifically to infer that they refer to objects not referred to by the query. 
This helps rank proposals; we use CPN to optimize using a reinforcement learning policy gradient algorithm.

We first present the framework of QRC Net, followed by the details of PGN, QRN and CPN respectively. Finally, we illustrate how to jointly optimize QRC Net and employ QRC Net in phrase grounding task.

\subsection{Framework}

The goal of QRC Net is to localize the mentioned object's location $y$ given an image $x$ and a query phrase $q$.
To achieve this, PGN generates a set of $N$ proposals $\{r_i\}$ as candidates.
Given the query $q$, QRN predicts their regression parameters $\{\mathbf{t}_i\}$ and probability $\{p_i\}$ of being relevant to the input query. 
To reduce semantic ambiguity, CPN evaluates prediction results of QRN based on the locations of objects mentioned by context phrases, and adopts a reward function $F$ to adaptively penalize high ranked proposals containing context-mentioned objects. 
Reward calculation depends on predicted proposals, and this procedure is non-differentiable. 
To overcome this, we deploy a reinforcement learning procedure in CPN where this reward is back propagated as policy gradients~\cite{sutton1999policy} to optimize QRN's parameters, which guides QRN to predict more discriminative proposals.
The objective for QRC Net is:
\begin{equation}\label{equ: obj of QRC Net}
\begin{aligned}
\arg\min_{\theta}\sum_q&\left[\mathcal{L}_{gen}(\{r_i\})\right. + \mathcal{L}_{cls}(\{r_i\}, \{p_i\}, y) \\
&\left. + \lambda\mathcal{L}_{reg}(\{r_i\}, \{\mathbf{t}_i\}, y)+J(\theta)\right]
\end{aligned}
\end{equation}
where $\theta$ denotes the QRC Net's parameters to be optimized and $\lambda$ is a hyperparameter. 
$\mathcal{L}_{gen}$ is the loss for generation proposals produced by PGN. 
$\mathcal{L}_{cls}$ is a multi-class classification loss generated by QRN in predicting the probability $p_i$ of each proposal $r_i$. 
$\mathcal{L}_{reg}$ is a regression loss from QRN to regress each proposal $r_i$ to the mentioned object's location $y$.
$J(\theta)$ is the reward expectation calculated by CPN.

\subsection{Proposal Generation Network (PGN)}\label{sec: pgn}
We build PGN with a similar structure as that of RPN in~\cite{ren2015faster}. 
PGN adopts a fully convolutional neural network (FCN) to encode the input image $x$ as an image feature map $\mathbf{x}$. 
For each location (\emph{i.e.}, anchor) in image feature map, PGN uses different scales and aspect ratios to generate proposals $\{r_i\}$.
Each anchor is fed into a multiple-layer perceptron (MLP) which predicts a probability $p_{o_i}$ estimating the objectness of the anchor, and 4D regression parameters $\mathbf{t}_i = [(x-x_a)/w_a, (y-y_a)/h_a, \log(w/w_a), \log(h/h_a)]$ as defined in~\cite{ren2015faster}.
The regression parameters $\mathbf{t}_i$ estimate the offset from anchor to mentioned objects' bounding boxes.
Given all mentioned objects' locations $\{y_l\}$, we consider a proposal to be positive when it covers some object $y_l$ with Intersection over Union (IoU) $>0.7$, and negative when IoU $< 0.3$. The generation loss is:
\begin{equation}\label{equ: L_gen}
\begin{aligned}
&\mathcal{L}_{gen}(\{r_i\}) = -\frac{1}{N_{cls}}\sum_{i=1}^{N_{cls}}\delta(i\in S_y\cup S_{\bar{y}})\log(p_{o_i})    \\
   & + \frac{\lambda_g}{N_{reg}}\sum_{i=1}^{N_{reg}}\delta(i\in S_y)\sum_{j=0}^3f\left(\left|\mathbf{t}_i^*[j]-\mathbf{t}_i[j]\right|\right)
\end{aligned}
\end{equation}
where $\delta(.)$ is an indicator function, $S_y$ is the set of positive proposals' indexes and $S_{\bar{y}}$ is the set of negative proposals' indexes. $N_{reg}$ is the number of all anchors and $N_{cls}$ is the number of sampled positive and negative anchors as defined in~\cite{ren2015faster}. $\mathbf{t}_i^*$ represents regression parameters of anchor $i$ to corresponding object's location $y_l$. 
$f(.)$ is the smooth L1 loss function: $f(x)=0.5x^2 (|x|<1)$, and $f(x) = |x|-0.5 (|x|\geq 1)$.

We sample the top $N$ anchors based on $\{p_{o_i}\}$ and regress them as proposals $\{r_i\}$ with predicted regression parameters $\mathbf{t}_i$. 
Through a RoI pooling operation~\cite{ren2015faster}, we extract visual feature $\mathbf{v}_i\in\mathbb{R}^{d_v}$ for each proposal $r_i$. $\{r_i\}$ and $\{\mathbf{v}_i\}$ as fed into QRN as visual inputs.

\subsection{Query guided Regression Network (QRN)}
For input query $q$, QRN encodes its semantics as an embedding vector $\mathbf{q}\in\mathbb{R}^{d_q}$ via a Long Short-Term Memory (LSTM) model.
Given visual inputs $\{\mathbf{v}_i\}$, QRN concatenates the embedding vector $\mathbf{q}$ with each of the proposal's visual feature $\mathbf{v}_i$. 
It then applies a fully-connected (fc) layer to generate multimodal features $\{\mathbf{v}_i^q\}\in\mathbb{R}^m$ for each of the $\langle q, r_i\rangle$ pair in an $m$-dimensional subspace. 
The multimodal feature $\mathbf{v}_i^q$ is calculated as:
\begin{equation}\label{equ: vis lang proj}
\mathbf{v}_i^q = \varphi(\mathbf{W}_m(\mathbf{q} || \mathbf{v}_i)+\mathbf{b}_m)
\end{equation}
where $\mathbf{W}_m\in\mathbb{R}^{(d_q+d_v)\times m}, \mathbf{b}_m\in\mathbb{R}^m$ are projection parameters. $\varphi(.)$ is a non-linear activation function. ``$||$'' denotes a concatenation operator.

Based on the multimodal feature $\mathbf{v}_i^q$, QRN predicts a 5D vector $\mathbf{s}_i^{p}\in\mathbb{R}^5$ via a fc layer for each proposal $r_i$ (superscript ``p'' denotes prediction).
\begin{equation}\label{equ: prediction}
\mathbf{s}_i^p = \mathbf{W}_s\mathbf{v}_i^q+\mathbf{b}_s
\end{equation}
where $\mathbf{W}_s\in\mathbb{R}^{m\times5}$ and $\mathbf{b}_s\in\mathbb{R}^5$ are projection weight and bias to be optimized.
The first element in $\mathbf{s}_i^p$ estimates the confidence of $r_i$ being related to input query $q$'s semantics. 
The next four elements are regression parameters which are in the same form as $\mathbf{t}_i$ defined in Sec.~\ref{sec: pgn}, where $x, y, w, h$ are replaced by regressed values and $x_a, y_a, w_a, h_a$ are proposal's parameters.

We denote $\{p_i\}$ as the probability distribution of $\{r_i\}$ after we feed $\{\mathbf{s}_i^p[0]\}$ to a softmax function.
Same as~\cite{rohrbach2016grounding}, we consider one proposal as positive which overlaps most with ground truth and with IoU $>0.5$. 
Thus, the classification loss is calculated as:
\begin{equation}\label{equ: L_cls_s}
\mathcal{L}_{cls}(\{r_i\}, \{p_i\}, y) = -\log(p_{i^*})
\end{equation}
where $i^*$ is positive proposal's index in the proposal set. 

Given the object's location $y$ mentioned by query $q$, each proposal's ground truth regression data $\mathbf{s}_i^q\in\mathbb{R}^4$ is calculated in the same way as the last four elements of $\mathbf{s}_i^p$, by replacing $[x,y,w,h]$ with the ground truth bounding box's location information.
The regression loss for QRN is:
\begin{equation}\label{equ: L_reg_s}
\mathcal{L}_{reg}(\{\mathbf{t}_i\}, \{r_i\}, y) = \frac{1}{4N}\sum_{i=1}^N\sum_{j=0}^3f\left(\left|\mathbf{s}_i^p[j+1]-\mathbf{s}_i^q[j]\right|\right)
\end{equation}
where $f(.)$ is the smooth L1 function defined in Sec.~\ref{sec: pgn}.

\subsection{Context Policy Network (CPN)}
Besides using QRN to predict and regress proposals, we further apply a CPN to guide QRN to avoid selecting proposals which cover the objects referred by query $q$'s context in the same description. CPN evaluates and assigns rewards for top ranked proposals produced from QRN, and performs a non-differentiable policy gradient~\cite{sutton1999policy} to update QRN's parameters.

Specifically, proposals $\{r_i\}$ from QRN are first ranked based on their probability distribution $\{p_i\}$.
Given the ranked proposals, CPN selects the top $K$ proposals $\{r_i^\prime\}$ and evaluates them by assigning rewards. 
This procedure is non-differentiable, since we do not know the proposals' qualities until they are ranked based on QRN's probabilities. 
Therefore, we use policy gradients reinforcement learning to update the QRN's parameters. 
The goal is to maximize the expectation of predicted reward $F(\{r_i^\prime\})$ under the distribution of $\{r_i^\prime\}$ parameterized by the QRN, \emph{i.e.}, $J = \mathbb{E}_{\{p_i\}}[F]$. 
According to the algorithm in~\cite{williams1992simple}, the policy gradient is
\begin{equation}\label{equ: policy grad}
\nabla_{\theta_r}J = \mathbb{E}_{\{p_i\}}[F(\{r_i^\prime\})\nabla_{\theta_r}\log p_i^\prime(\theta_r)]
\end{equation}
where $\theta_r$ are QRN's parameters and $\nabla_{\theta_r}\log p_i^\prime(\theta_r)$ is the gradient produced by QRN for top ranked proposal $r_i$.

To predict reward value $F(\{r_i^\prime\})$, CPN averages top ranked proposals' visual features $\{\mathbf{v}_i^\prime\}$ as $\mathbf{v}_c$. 
The predicted reward is computed as:
\begin{equation}\label{equ: reward}
F(\{r_i^\prime\}) = \sigma(\mathbf{W}_c(\mathbf{v}_c||\mathbf{q})+\mathbf{b}_c)
\end{equation}
where ``$||$'' denotes concatenation operation and $\sigma$(.) is a sigmoid function. 
$\mathbf{W}_c$ and $\mathbf{b}_c$ are projection parameters which produce a scalar value as reward.

To train CPN, we design a reward function to guide CPN's prediction. The reward function performs as feedback from environment and guide CPN to produce meaningful policy gradients.
Intuitively, to help QRN select more discriminative proposals related to query $q$ rather than context, we assign lower reward for some top ranked proposal that overlaps the object mentioned by context and higher reward if it overlaps with the mentioned object by query.
Therefore, we design the reward function as:
\begin{equation}\label{equ: rwd func}
R(\{r_i^\prime\}) = \frac{1}{K}\sum_{i=1}^K[\delta(r_i^\prime\in S_q) + \beta\delta(r_i^\prime\notin(S_q\cup S_{bg}))]
\end{equation}
where $S_q$ is the set of proposals with IoU $>0.5$ with mentioned objects by query $q$, and $S_{bg}$ is the set of background proposals with IoU $<0.5$ with objects mentioned by all queries in the description. $\delta$(.) is an indicator function and $\beta\in(0, 1)$ is the reward for proposals overlapping with objects mentioned by context.
The reward prediction loss is:
\begin{equation}\label{equ: rwd loss}
\mathcal{L}_{rwd}(\{r_i^\prime\}) = ||F(\{r_i^\prime\})-R(\{r_i^\prime\})||^2
\end{equation}
During training, $\mathcal{L}_{rwd}$ is backpropagated only to CPN for optimization, while CPN backpropagates policy gradients (Eq.~\ref{equ: policy grad}) to optimize QRN.


\subsection{Training and Inference}
We train PGN based on an RPN pre-trained on PASCAL VOC 2007~\cite{pascal-voc-2007} dataset, and adopt the alternating training method in~\cite{ren2015faster} to optimize PGN. 
We first train PGN and use proposals to train QRN and CPN, then initialize PGN tuned by QRN and CPN's training, which iterates one time. 
Same as~\cite{rohrbach2016grounding}, we select 100 proposals produced by PGN ($N=100$) and select top 10 proposals ($K=10$) predicted by QRN to assign reward in Eq.~\ref{equ: rwd func}.
After calculating policy gradient in Eq.~\ref{equ: policy grad}, we jointly optimize QRC Net's objective (Eq.~\ref{equ: obj of QRC Net}) using Adam algorithm~\cite{kingma2014adam}. We choose the rectified linear unit (ReLU) as the non-linear activation function. 

During testing stage, CPN is fixed and we stop its reward calculation. 
Given an image, PGN is first applied to generate proposals and their visual features. QRN regresses these proposals and predicts the relevance of each proposal to the query. The regressed proposal with highest relevance is selected as the prediction result.

\section{Experiment}\label{sec: exps}
We evaluate QRC Net on Flickr30K Entities~\cite{plummer2015flickr30k} and Referit Game datasets~\cite{KazemzadehOrdonezMattenBergEMNLP14} for phrase grounding task.

\subsection{Datasets}
\textbf{Flickr30K Entities}~\cite{plummer2015flickr30k}: 
The numbers of training, validation and testing images are 29783, 1000, 1000 respectively.
Each image is associated with 5 captions, with 3.52 query phrases in each caption on average.
There are 276K manually annotated bounding boxes referred by 360K query phrases in images.
The vocabulary size for all these queries is 17150.

\textbf{Referit Game}~\cite{KazemzadehOrdonezMattenBergEMNLP14} consists of 19,894 images of natural scenes. 
There are 96,654 distinct objects in these images. 
Each object is referred to by 1-3 query phrases (130,525 in total).
There are 8800 unique words among all the phrases, with a maximum length of 19 words.

\subsection{Experiment Setup}\label{sec: exp setup}
\textbf{Proposal generation.} We adopt a PGN (Sec.~\ref{sec: pgn}) to generate proposals. During training, we optimize PGN based on an RPN pre-trained on PASCAL VOC 2007 dataset~\cite{pascal-voc-2007}, which does not overlap with Flickr30K Entities~\cite{plummer2015flickr30k} or Referit Game~\cite{KazemzadehOrdonezMattenBergEMNLP14}. We also evaluate QRC Net based on Selective Search~\cite{uijlings2013selective} (denoted as ``SS'') and EdgeBoxes~\cite{zitnick2014edge} (denoted as ``EB''), and an RPN~\cite{ren2015faster} pre-trained on PASCAL VOC 2007~\cite{pham2013fast} (denoted as ``RPN''), which are all independent of QRN and CPN.

\textbf{Visual feature representation.}
For QRN, the visual features are directly generated from PGN via a RoI pooling operation. Since PGN contains a VGG Network~\cite{Simonyan14c} to process images, we denote these features as ``VGG\textsubscript{pgn}''. 
To predict regression parameters, we need to include spatial information for each proposal. 
For Flickr30K Entities, we augment each proposal's visual feature with its spatial information $[x_{tl}/W, y_{tl}/H, x_{br}/H, y_{br}/W, wh/WH]$ as defined in~\cite{yu2016modeling}. These augmented features are 4101D vectors ($d_v=4101$).
For Referit Game, we augment VGG\textsubscript{pgn} with each proposal's spatial information $[x_{min},y_{min},x_{max},y_{max},$ $x_{center}, y_{center}, w_{box}, h_{box}]$ which is same as~\cite{rohrbach2016grounding} for fair comparison. 
We denote these features as ``VGG\textsubscript{pgn}-SPAT'', which are 4104D vectors ($d_v=4104$).

To compare with other approaches, we replace PGN with a Selective Search and an EdgeBoxes proposal generator. 
Same as~\cite{rohrbach2016grounding}, we choose a VGG network finetuned using Fast-RCNN~\cite{girshickICCV15fastrcnn} on PASCAL VOC 2007~\cite{pascal-voc-2007} to extract visual features for Flickr30K Entities. We denote these features as ``VGG\textsubscript{det}''.
Besides, we follow~\cite{rohrbach2016grounding} and apply a VGG network pre-trained on ImageNet~\cite{deng2009imagenet} to extract proposals' features for Flickr30K Entities and Referit Game, which are denoted as ``VGG\textsubscript{cls}''.
We augment VGG\textsubscript{det} and VGG\textsubscript{cls} with spatial information for Flickr30K Entities and Referit Game datasets following the method mentioned above. 

\textbf{Model initialization.}
Following same settings as in~\cite{rohrbach2016grounding}, we encode queries via an LSTM model, and choose the last hidden state from LSTM as $\mathbf{q}$ (dimension $d_q=1000$).
All convolutional layers are initialized by MSRA method~\cite{he2015delving} and all fc layers are initialized by Xavier method~\cite{glorot2010understanding}.
We introduce batch normalization layers after projecting visual and language features (Eq.~\ref{equ: vis lang proj}).

During training, the batch size is 40. We set weight $\lambda$ for regression loss $L_{reg}$ as 1.0 (Eq.~\ref{equ: obj of QRC Net}), and reward value $\beta=0.2$ (Eq.~\ref{equ: rwd func}). The dimension of multimodal feature vector $\mathbf{v}_i^q$ is set to $m=512$ (Eq.~\ref{equ: vis lang proj}). Analysis of hyperparameters is provided in Sec.~\ref{sec: flickr perf} and~\ref{sec: referit perf}.

\textbf{Metric.} Same as~\cite{rohrbach2016grounding}, we adopt accuracy as the evaluation metric, defined to be the ratio of phrases for which the regressed box overlaps with the mentioned object by more than 50\% IoU.

\textbf{Compared approaches.}~We choose GroundeR~\cite{rohrbach2016grounding}, CCA embedding~\cite{plummer2015flickr30k}, MCB~\cite{fukui2016multimodal}, Structured Matching~\cite{wang2016structured} and SCRC~\cite{hu2016natural} for comparison, which all achieve leading performances in phrase grounding. 
For GroundeR~\cite{rohrbach2016grounding}, we compare with its supervised learning scenario, which achieves the best performance among different scenarios.

\begin{table}[t]
  \centering
  \begin{tabular}{lc} \toprule
  Approach & Accuracy (\%) \\ \midrule
  \textbf{Compared approaches} & \\
  SCRC~\cite{hu2016natural} & 27.80 \\
  Structured Matching~\cite{wang2016structured} & 42.08 \\
  SS+GroundeR (VGG\textsubscript{cls})~\cite{rohrbach2016grounding} & 41.56 \\
  RPN+GroundeR (VGG\textsubscript{det})~\cite{rohrbach2016grounding} & 39.13 \\
  SS+GroundeR (VGG\textsubscript{det})~\cite{rohrbach2016grounding} & 47.81 \\
  MCB~\cite{fukui2016multimodal} & 48.69 \\
  CCA embedding~\cite{plummer2015flickr30k} & 50.89 \\ \midrule
  \textbf{Our approaches} & \\
  RPN+QRN (VGG\textsubscript{det})  & 53.48 \\
  SS+QRN (VGG\textsubscript{det}) & 55.99 \\     
  PGN+QRN (VGG\textsubscript{pgn}) & 60.21 \\ 
  QRC Net (VGG\textsubscript{pgn})& \textbf{65.14} \\
  \bottomrule
  \end{tabular}
  \vspace{1.0mm}
\caption{Different models' performance on Flickr30K Entities. Our framework is evaluated by combining with various proposal generation systems.}\label{tab: flickr30k res}
\end{table}

\begin{table}[t]
  \centering
  \begin{tabular}{|l|c|c|c|} \hline
  Proposal generation & RPN~\cite{ren2015faster} & SS~\cite{uijlings2013selective} & PGN \\ \hline
  UBP (\%) & 71.25 & 77.90 & 89.61 \\ \hline
  BPG & 7.29 & 3.62 & 7.53 
  \\\hline 
  \end{tabular}
  \vspace{1.0mm}
\caption{Comparison of different proposal generation systems on Flickr30k Entities}\label{tab: prop comparison}
\end{table}

\subsection{Performance on Flickr30K Entities}\label{sec: flickr perf}
\textbf{Comparison in accuracy.} We first evaluate QRN performance based on different independent proposal generation systems. 
As shown in Table~\ref{tab: flickr30k res}, by adopting QRN, RPN+QRN achieves 14.35\% increase compared to RPN+GorundeR.
We further improve QRN's performance by adopting Selective Search (SS) proposal generator. 
Compared to SS+GroundeR, we achieve 8.18\% increase in accuracy.
We then incorporate our own PGN into the framework, which is jointly optimized to generate proposals as well as features (VGG\textsubscript{pgn}).
By adopting PGN, PGN+QRN achieves 4.22\% increase compared to independent proposal generation system (SS+QRN) in accuracy.
Finally, we include CPN to guide QRN in selecting more discriminative proposals during training. 
The full model (QRC Net) achieves 4.93\% increase compared to PGN+QRN, and 14.25\% increase over the state-of-the-art CCA embedding~\cite{plummer2015flickr30k} in accuracy.

\begin{table}[t]
  \centering
  \begin{tabular}{|l|c|c|c|c|c|} \hline
  Weight $\lambda$ & 0.5 & 1.0 & 2.0 & 4.0 & 10.0 \\ \hline
  Accuracy (\%) & 64.15 & 65.14 & 64.40 & 64.29 & 63.27
  \\\hline 
  \end{tabular}
  \vspace{1.0mm}
\caption{QRC Net's performances on Flickr30K Entities for different weights $\lambda$ of $L_{reg}$.}\label{tab: lbd flickr}
\end{table}

\begin{table}[t]
  \centering
  \begin{tabular}{|l|c|c|c|c|} \hline
  Dimension $m$ & 128 & 256 & 512 & 1024 \\ \hline
  Accuracy (\%) & 64.08 & 64.59 & 65.14 & 62.52 
  \\\hline 
  \end{tabular}
  \vspace{1.0mm}
\caption{QRC Net's performances on Flickr30K Entities for different dimensions $m$ of $\mathbf{v}_i^q$.}\label{tab: m flickr}
\end{table}

\begin{table}[t]
  \centering
  \begin{tabular}{|l|c|c|c|c|} \hline
  Reward $\beta$ & 0.1 & 0.2 & 0.4 & 0.8 \\ \hline
  Accuracy (\%) & 64.10 & 65.14 & 63.88 & 62.77  
  \\\hline 
  \end{tabular}
  \vspace{1.0mm}
\caption{QRC Net's performances on Flickr30K Entities for different reward values $\beta$ of CPN.}\label{tab: beta flickr}
\end{table}

\textbf{Detailed comparison.} Table~\ref{tab: flickr30k detail res} provides the detailed phrase localization results based on the phrase type information for each query in Flickr30K Entities. 
We can observe that QRC Net provides consistently superior results.
CCA embedding~\cite{plummer2015flickr30k} model is good at localizing ``instruments'' while GroundeR~\cite{rohrbach2016grounding} is strong in localizing ``scene''. 
By using QRN, we observe that the regression network achieves consistent increase in accuracy compared to GroundeR model (VGG\textsubscript{det}) in all phrase types except for class ``instruments''. 
Typically, there is a large increase in performance of localizing ``animals'' (with increase of 11.39\%).
By using PGN, we observe that PGN+QRN has surpassed state-of-the-art method in all classes, with largest increase in class ``instruments''.   
Finally, by applying CPN, QRC Net achieves more than 8.03\%, 9.37\%, 8.94\% increase in accuracy in all categories compared to CCA embedding~\cite{plummer2015flickr30k}, Structured Matching~\cite{wang2016structured} and GroundeR~\cite{rohrbach2016grounding} respectively.
QRC Net achieves the maximum increase in performance of 15.73\% for CCA embedding~\cite{plummer2015flickr30k} (``scene''), 32.90\% for Structured Matching~\cite{wang2016structured} (``scene'') and  21.46\% for GroundeR~\cite{rohrbach2016grounding} (``clothing''). 

\begin{figure*}[ht]
\includegraphics[width=6.5in]{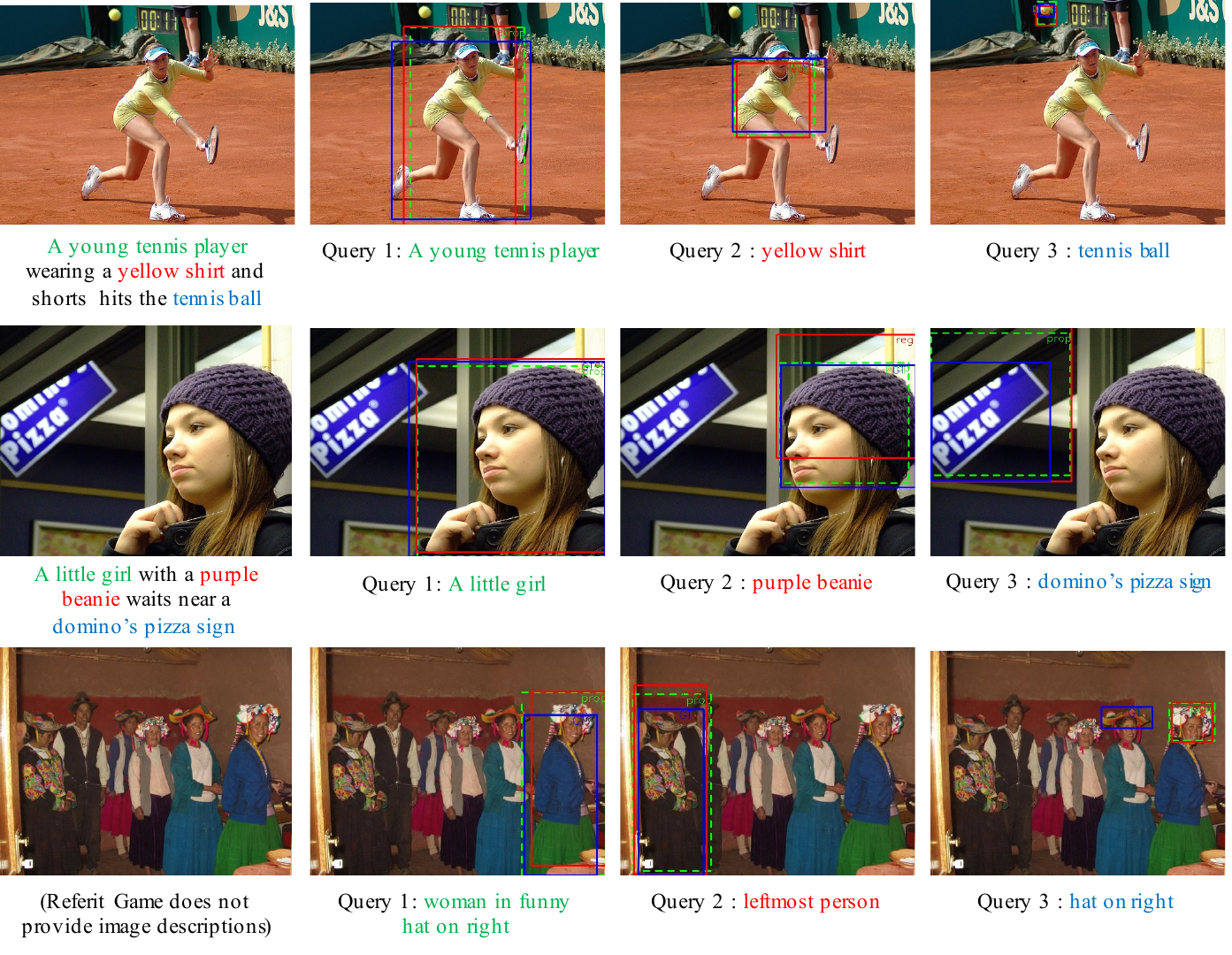}
\centering
\caption{Some phrase grounding results in Flickr30K Entities~\cite{plummer2015flickr30k} (first two rows) and Referit Game~\cite{KazemzadehOrdonezMattenBergEMNLP14} (third row). We visualize ground truth bounding box, selected proposal box and regressed bounding box in blue, green and red resepctively. When query is not clear without further context information, QRC Net may ground wrong objects (\emph{e.g.}, image in row three, column four). }\label{fig: demo flickr}
\end{figure*}

\begin{table*}[t]
  \centering
  \begin{tabular}{lcccccccc} \toprule
  Phrase Type & people & clothing & body parts & animals & vehicles & instruments & scene & other  \\ \midrule
  GroundeR (VGG\textsubscript{cls})~\cite{rohrbach2016grounding} & 53.80 & 34.04 & 7.27 & 49.23 & 58.75 & 22.84 & 52.07 & 24.13 \\
  GroundeR (VGG\textsubscript{det})~\cite{rohrbach2016grounding} & 61.00 & 38.12 & 10.33 & 62.55 & 68.75 & 36.42 & 58.18 & 29.08 \\
  Structured Matching~\cite{wang2016structured} & 57.89 & 34.61 & 15.87 & 55.98 & 52.25 & 23.46 & 34.22 & 26.23 \\
  CCA embedding~\cite{plummer2015flickr30k} & 64.73 & 46.88 & 17.21 & 65.83 & 68.75 & 37.65 & 51.39 & 31.77 \\ \midrule
  SS+QRN  & 68.24 & 47.98 & 20.11 & 73.94 & 73.66 & 29.34 & 66.00 & 38.32 \\
  PGN+QRN & 75.08 & 55.90 & 20.27 & 73.36 & 68.95 & 45.68 & 65.27 & 38.80 \\
  QRC Net & \textbf{76.32} & \textbf{59.58} & \textbf{25.24} & \textbf{80.50} & \textbf{78.25} & \textbf{50.62} & \textbf{67.12} & \textbf{43.60} \\\bottomrule 
  \end{tabular}
  \vspace{1.0mm}
\caption{Phrase grounding performances for different phrase types defined in Flickr30K Entities. Accuracy is in percentage.}\label{tab: flickr30k detail res}\label{tab: flickr30k detail res}
\end{table*}

\textbf{Proposal generation comparison.} We observe proposals' quality plays an important role in final grounding performance. 
The influence has two aspects. 
First is the Upper Bound Performance (UBP) which is defined as the ratio of covered objects by generated proposals in all ground truth objects. 
Without regression mechanism, UBP directly determines the performance limit of grounding systems. 
Another aspect is the average number of surrounding Bounding boxes Per Ground truth object (BPG). Generally, when BPG increases, more candidates are considered as positive, which reduces the difficulty for following grounding system. 
To evaluate UBP and BPG, we consider that a proposal covers the ground truth object when its IoU $>0.5$. The statistics for RPN, SS and PGN in these two aspects are provided in Table~\ref{tab: prop comparison}. 
We observe that PGN achieves increase in both UBP and PBG, which indicates PGN provides high quality proposals for QRN and CPN.
Moreover, since QRN adopts a regression-based method, it can surpass UBP of PGN, which further relieves the influence from UBP of proposal generation systems.

\textbf{Hyperparameters.} We evaluate QRC Net for different sets of hyperparameters. 
To evaluate one hyperparameter, we fix other hyperparameters to default values in Sec.~\ref{sec: exp setup}.

We first evaluate QRC Net's performance for different regression loss weights $\lambda$. The results are shown in Table~\ref{tab: lbd flickr}. 
We observe the performance of QRC Net fluctuates when $\lambda$ is small and decreases when $\lambda$ becomes large.

We then evaluate QRC Net's performance for different dimensions $m$ for multimodal features in Eq.~\ref{equ: vis lang proj}. The performances are presented in Table~\ref{tab: m flickr}. 
We observe QRC Net's performance fluctuates when $m < 1000$. When $m$ becomes large, the performance of QRC Net decreases.
Basically, these changes are in a small scale, which shows the insensitivity of QRC Net to these hyperparameters.

Finally, we evaluate different reward values $\beta$ for proposals covering objects mentioned by context. We observe QRC Net's performance fluctuates when $\beta<0.5$. When $\beta$ is close to 1.0, the CPN assigns almost same rewards for proposals covering ground truth objects or context mentioned objects, which confuses the QRN. As a result, the performance of QRC Net decreases.

\subsection{Performance on Referit Game}\label{sec: referit perf}
\begin{table}[t]
  \centering
  \begin{tabular}{lc} \toprule
  Approach & Accuracy (\%) \\ \midrule
  \textbf{Compared approaches} & \\
  SCRC~\cite{hu2016natural} & 17.93 \\
  EB+GroundeR (VGG\textsubscript{cls}-SPAT)~\cite{rohrbach2016grounding} & 26.93 \\ \midrule
  \textbf{Our approaches} & \\
  EB+QRN (VGG\textsubscript{cls}-SPAT) & 32.21 \\
  PGN+QRN (VGG\textsubscript{pgn}-SPAT) & 43.57 \\
  QRC Net (VGG\textsubscript{pgn}-SPAT) & \textbf{44.07} \\
  \bottomrule
  \end{tabular}
  \vspace{1.0mm}
\caption{Different models' performance on Referit Game dataset.}\label{tab: referit res}
\end{table}
\textbf{Comparison in accuracy.} 
To evaluate QRN's effectiveness, we first adopt an independent EdgeBoxes~\cite{zitnick2014edge} (EB) as proposal generator, which is same as~\cite{rohrbach2016grounding}. 
As shown in Table~\ref{tab: referit res}, by applying QRN, we achieve 5.28\% improvement compared to EB+GroundeR model. 
We further incorporate PGN into the framework. 
PGN+QRN model brings 11.36\% increase in accuracy, which shows the high quality of proposals produced by PGN.
Finally, we evaluate the full QRC Net model. Since Referit Game dataset only contains independent query phrases, there is no context information available. 
In this case, only the first term in Eq.~\ref{equ: rwd func} guides the learning. Thus, CPN does not contribute much to performance (0.50\% increase in accuracy).

\textbf{Hyperparameters.} We evaluate QRC Net's performances for different hyperparameters on Referit Game dataset. 
First, we evaluate QRC Net's performance for different weights $\lambda$ of regression loss $L_{reg}$. 
As shown in Table~\ref{tab: lbd referit}, performance of QRC Net fluctuates when $\lambda$ is small. When $\lambda$ becomes large, regression loss overweights classification loss, where a wrong seed proposal may be selected which produces wrong grounding results. Thus, the performance decreases.

\begin{table}[t]
  \centering
  \begin{tabular}{|l|c|c|c|c|c|} \hline
  Weight $\lambda$ & 0.5 & 1.0 & 2.0 & 4.0 & 10.0 \\ \hline
  Accuracy (\%) & 43.71 & 44.07 & 43.61 & 43.60 & 42.75
  \\\hline 
  \end{tabular}
  \vspace{1.0mm}
\caption{QRC Net's performances on Referit Game for different weights $\lambda$ of $L_{reg}$.}\label{tab: lbd referit}
\end{table}

\begin{table}[t]
  \centering
  \begin{tabular}{|l|c|c|c|c|} \hline
  Dimension $m$ & 128 & 256 & 512 & 1024 \\ \hline
  Accuracy (\%) & 42.95 & 43.80 & 44.07 & 43.51  
  \\\hline 
  \end{tabular}
  \vspace{1.0mm}
\caption{QRC Net's performances on Regerit Game for different dimensions $m$ of $\mathbf{v}_i^q$.}\label{tab: m referit}
\end{table}

We then evaluate QRC Net's performance for different multimodal dimensions $m$ of $\mathbf{v}_i^q$ in Eq.~\ref{equ: vis lang proj}. 
In Table~\ref{tab: m referit}, we observe performance changes in a small scale when $m<1000$, and decreases when $m>1000$.

\subsection{Qualitative Results}
We visualize some phrase grounding results of Flickr30K Entities and Referit Game for qualitative evaluation (Fig.~\ref{fig: demo flickr}). 
For Flickr30K Entities, we show an image with its associated caption, and highlight the query phrases in it. 
For each query, we visualize the ground truth box, the selected proposal box by QRN and the regressed bounding box based on the regression parameters predicted by QRN. 
Since there is no context information in Referit Game, we visualize query and ground truth box, with selected proposal and regressed box predicted by QRN. 

As shown in Fig~\ref{fig: demo flickr}, QRC Net is strong in recognizing different people (``A young tennis player'' in the first row) and clothes (``purple beanie'' in the second row), which is also validated in Table~\ref{tab: flickr30k detail res}. 
However, when the query is ambiguous without further context description, QRC Net may be confused and produce reasonably incorrect grounding result (\emph{e.g.}, ``hat on the right'' in the third row of Fig.~\ref{fig: demo flickr}).

\section{Conclusion}
We proposed a novel deep learning network (QRC Net) to address the phrase grounding task. 
QRC Net adopts regression mechanism and leverages context information, which achieves 14.25\% and 17.14\% increase in accuracy on Flickr30K Entities~\cite{plummer2015flickr30k} and Referit Game~\cite{KazemzadehOrdonezMattenBergEMNLP14} datasets respectively.

\section*{Acknowledgements}
This paper is based, in part, on research sponsored by the Air Force Research Laboratory and the Defense Advanced Research Projects Agency under agreement number FA8750-16-2-0204. 
The U.S. Government is authorized to reproduce and distribute reprints for Governmental purposes notwithstanding any copyright notation thereon. The views and conclusions contained herein are those of the authors and should not be interpreted as necessarily representing the official policies or endorsements, either expressed or implied, of the Air Force Research Laboratory and the Defense Advanced Research Projects Agency or the U.S. Government.

{\small
\bibliographystyle{ieee}
\bibliography{egbib}
}

\end{document}